\documentclass[conference]{IEEEtran}
\IEEEoverridecommandlockouts
\usepackage{booktabs} 
\usepackage{graphicx}
\usepackage{xcolor,colortbl}
\definecolor{storeClusterComponent}{HTML}{FDAE61}
\definecolor{dbscan}{HTML}{ABDDA4}
\usepackage{csquotes}
\usepackage{amsmath}
\usepackage{amssymb}
\usepackage{amsfonts}
\usepackage{url}
\usepackage{hyphenat}
\usepackage{epsfig}
\usepackage{color}
\usepackage{makecell}
\usepackage[font={small,it}]{caption}
\usepackage{array,multirow}
\usepackage{epstopdf}
\usepackage{tabularx}
\usepackage{subfig,stfloats}
\usepackage{ifthen}
\usepackage{multirow}
\usepackage{booktabs}
\usepackage{lipsum}
\usepackage{balance}
\usepackage{bbm}
\usepackage{accents}
\usepackage{hyperref}
\usepackage{tikz,pgfplots,pgfplotstable}
\usetikzlibrary{pgfplots.groupplots}
\usetikzlibrary{positioning, arrows, shapes}
\usepackage{algorithm}
\usepackage{soul}
\usepackage[noend]{algpseudocode}

\def\BibTeX{{\rm B\kern-.05em{\sc i\kern-.025em b}\kern-.08em
    T\kern-.1667em\lower.7ex\hbox{E}\kern-.125emX}}

\usepackage{booktabs} 
\pgfplotsset{compat=1.11,
    /pgfplots/ybar legend/.style={
    /pgfplots/legend image code/.code={%
       \draw[##1,/tikz/.cd,yshift=-0.25em]
        (0cm,0cm) rectangle (0.8em,4pt);},
   },
}

\newtheorem{examp}{Example}
\newcommand{\nop}[1]{}

\def\G{\mathcal G}

\usepackage{anyfontsize}

\begin{document}

\title{{Multi-Stage Graph Peeling Algorithm for Probabilistic Core Decomposition}
\thanks{This work is supported in part by funds from the Natural Sciences and Engineering Research Council of Canada: NSERC Discovery Grants: \# RGPIN-2017-04722 (Y.G. \& X.Z.), \# RGPIN-2017-04039 (V.S.), \# RGPIN-2016-04022 (A.T.), \# RGPIN-2021-03530 (L.X.) and the Canada Research Chair \#950-231363 (X.Z.).}
}

\author{\IEEEauthorblockN{Yang Guo}
\IEEEauthorblockA{\textit{Dept. of Mathematics and Statistics} \\
\textit{University of Victoria}\\
Victoria, BC, V8W 2Y2, Canada \\
yangguo@uvic.ca}
\and
\IEEEauthorblockN{Xuekui Zhang}
\IEEEauthorblockA{\textit{Dept. of Mathematics and Statistics} \\
\textit{University of Victoria}\\
Victoria, BC, V8W 2Y2, Canada \\
xuekui@uvic.ca}
\and
\IEEEauthorblockN{Fatemeh Esfahani}
\IEEEauthorblockA{\textit{Dept. of Computer Science} \\
\textit{University of Victoria}\\
Victoria, BC, V8W 2Y2, Canada \\
esfahani@uvic.ca}
\and
\IEEEauthorblockN{Venkatesh Srinivasan}
\IEEEauthorblockA{\textit{Dept. of Computer Science} \\
\textit{University of Victoria}\\
Victoria, BC, V8W 2Y2, Canada \\
srinivas@uvic.ca}
\and
\IEEEauthorblockN{Alex Thomo}
\IEEEauthorblockA{\textit{Dept. of Computer Science} \\
\textit{University of Victoria}\\
Victoria, BC, V8W 2Y2, Canada \\
thomo@uvic.ca}
\and
\IEEEauthorblockN{Li Xing}
\IEEEauthorblockA{\textit{Dept. of Mathematics and Statistics} \\
\textit{University of Saskatchewan}\\
Saskatoon, SK, S7N 5A2, Canada \\
lix491@mail.usask.ca}
}

\maketitle

\begin{abstract}
Mining dense subgraphs where vertices connect closely with each other is a common task when analyzing graphs. A very popular notion in subgraph analysis is core decomposition. Recently, Esfahani \emph{et al.} presented a probabilistic core decomposition algorithm based on graph peeling and Central Limit Theorem (CLT) that is capable of handling very large graphs. Their proposed peeling algorithm (PA) starts from the lowest degree vertices and recursively deletes these vertices, assigning core numbers, and updating the degree of neighbour vertices until it reached the maximum core. However, in many applications, particularly in biology, more valuable information can be obtained from dense sub-communities and we are not interested in small cores where vertices do not interact much with others. To make the previous PA focus more on dense subgraphs, we propose a multi-stage graph peeling algorithm (M-PA) that has a two-stage data screening procedure added before the previous PA. After removing vertices from the graph based on the user-defined thresholds, we can reduce the graph complexity largely and without affecting the vertices in subgraphs that we are interested in. We show that M-PA is more efficient than the previous PA and with the properly set filtering threshold, can produce very similar if not identical dense subgraphs to the previous PA (in terms of graph density and clustering coefficient).
\end{abstract}


\section{Introduction}
Dense subgraph mining is a fundamental task in many graph analytic tasks. Studying dense subgraphs can reveal important information about connectivity, centrality, and robustness of the network. For instance, we can find sub-community of users with close relationships in social networks; locate highly active pathways in gene expression networks, and identify clusters of suspicious accounts with possible money laundering behaviours from networks of transaction histories. There exist different definitions of dense subgraphs such as $k$-cliques, $k$-plexes, and $n$-clubs, which are intractable to compute~\cite{bonchi2014core}. Core decomposition is a popular notion of dense subgraphs due to its cohesive structure and the fact that it can be computed in polynomial time. It can also be used to compute other dense subgraphs such as maximal cliques~\cite{bonchi2014core}. The $k$-core is defined as the largest subgraph in which each vertex has a degree of at least $k$ within the subgraph. The collection of all $k$-cores for different values of $k$ forms the core decomposition of the graph. The highest value of $k$ for which a vertex belongs to a $k$-core subgraph is called the core number (or coreness) of the vertex. Core decomposition has been used in several applications such as text summarization~\cite{antiqueira2009complex}, exploring collaboration in software teams~\cite{wolf2009mining}, and describing biological functions of proteins in protein-protein interaction networks \cite{li2010computational}.

Extension of the definition of core decomposition to probabilistic graphs has been recently introduced in the literature~\cite{bonchi2014core}. Due to intrinsic uncertainty in many real-world networks such as social, biological, and communication networks (cf.~\cite{cheng2015reachability}) it is important to study core decomposition in probabilistic contexts. Probabilistic graphs are graphs in which each edge is assigned a probability of existence. In social and trust networks, an edge can be weighted by the probability of influence or trust between two users that the edge connects~\cite{korovaiko2013trust}. In biological networks of protein-protein interactions (cf.~\cite{Genome}), an edge can be assigned a probability value representing the strength of prediction that a pair of proteins will interact in a living organism~\cite{sharan2007network}.

We use the notion of $(k,\eta)$-core introduced by Bonchi \emph{et al.}~\cite{bonchi2014core}. Specifically, we aim to compute the largest subgraph in which each vertex has at least $k$ neighbours within that subgraph with probability no less than a user-specified threshold $\eta$. To compute core decomposition in probabilistic graphs, the $\eta$-degree or probabilistic degree of a vertex is introduced in~\cite{bonchi2014core}.  
The standard approach for computing $(k,\eta)$-core decomposition is the peeling process which is based on continuously removing the vertices with $\eta$-degree less than $k$~\cite{bonchi2014core,esfahani2019efficient}. When a vertex is removed, its core number is set to be its $\eta$-degree at the time of removal, and the $\eta$-degree of all its neighbours is computed and updated again. The peeling process is repeated after incrementing $k$ until no vertices remain, which results in finding all $(k,\eta)$-cores for different values of $k$, and user-defined threshold $\eta$. Core decomposition in probabilistic graphs is challenging due to the combinatorial nature of $\eta$-degree computation in such graphs.  Esfahani \emph{et al.}~\cite{esfahani2019efficient} improved the peeling process by using easy-to-compute lower-bounds on $\eta$-degree of vertices based on Lyapunov Central Limit Theorem (CLT) in statistics~\cite{lyapunov-clt} and designing efficient array structures for storing important bookkeeping information~\cite{esfahani2019efficient}. 

However, the previously proposed graph peeling algorithm by Esfahani \emph{et al.}~\cite{esfahani2019efficient} still works by starting from the lowest degree vertices and spends a lot of time working its way up to vertices that are more valuable in terms of information abundance. Our motivation is that comparing to low incidence vertices in the small cores, we think more valuable information can be found by directly focusing on more dense cores with high activities. 

In this work, we will present a more efficient and dense core focusing multi-stage peeling algorithm. Since it will be based on our previously developed algorithm, we will refer to the previous algorithm \textit{PA} and our proposed algorithm \textit{M-PA} in the rest of the paper. For M-PA, a two-stage filtering procedure is added in order for PA to properly screen out vertices in smaller cores and focus on denser sub-communities. The idea is after filtration, we have effectively decomposed the large graph into smaller subgraphs and we can then perform core decomposition even more efficiently.

\section{Background}

Let $G=(V,E)$ be an undirected graph, where $V$ and $E$ are the set of vertices and edges in $G$, respectively. Given a vertex $u \in V$, let $N_{G}(u)$ be the set of all neighbours of $u$, i.e. $N_{G}(u) = \{v: (u,v) \in E\}$. $\left| N_{G}(u) \right|$, is equal to deterministic degree of $u$ in $G$.

\subsection{Core Decomposition in Deterministic Graphs} 
Given a graph $G$, the $k$-core of $G$ is defined as the largest subgraph $H \subseteq G$ in which each vertex has degree of at least $k$ in $H$. The set of all $k$-cores forms the core decomposition of $G$, where $ 0 \leq k \leq d_{\max}(G)$, and $d_{\max}(G)$ is the maximum vertex degree in $G$. Given a vertex $u$, the largest value of $k$ for which $u$ belongs to a $k$-core is called core number of $u$.

\subsection{Probabilistic Graphs} 
A probabilistic graph $\G = (V, E, p)$, is defined over a set of vertices $V$, a set of edges $E$ and a probability function $p : E \rightarrow (0,1]$ which assigns an existence probability $p(e)$ to every edge $e \in E$. In the literature, the existence probability of each edge is assumed to be independent of other edges~\cite{bonchi2014core}. 

The \textit{possible worlds} of $\mathcal{G}$ are deterministic graph instances of $\G$, which are used for analyzing probabilistic graphs. In each possible world, only a subset of edges appears. 
For each possible world $G = (V, E_G) \sqsubseteq  \G $, where $E_G \subseteq E$, the probability of observing that possible world is obtained as follows:
$
    \text{Pr}(G) = \prod_{e \in E_G} p(e) \prod_{e \in E\setminus E_G}(1-p(e)).
$

\begin{figure}[H]
    \centering
    \subfloat[]{ \label{probgraph1}
     \begin{tikzpicture}[auto, node distance=2cm, every loop/.style={},
                    thick,main node/.style={scale=0.6, 
                    circle,draw,font=\sffamily\small\bfseries}]
                    \node[main node,fill={orange}] (1) {\textcolor{white}{1}};
                    \node[main node,fill={orange}] (2) [right of = 1] {\textcolor{white}{2}};
                     \node[main node,fill={orange}] (3) [below of = 1] {\textcolor{white}{3}};
                    \node[main node,fill={orange}] (4) [below of = 2] {\textcolor{white}{4}};
                     \node[main node,fill={green!70!blue}] (0) [below left of = 1] {\textcolor{white}{0}};
                     \node[main node,fill={green!70!blue}] (5) [above right of = 4] {\textcolor{white}{5}};
                      \path[every node/.style={scale=0.6,
                      font=\sffamily\small}]
                      (0) edge[] node [left, pos=0.6, font=\small\bfseries] {0.3} (1)
                      
                      (1) edge[] node [above, font=\small\bfseries] {0.4} (2)
                       edge[] node [pos = 0.5, left, font=\small\bfseries] {0.6} (3)
                        (2) edge[] node [right,pos=0.5, font=\small\bfseries] {0.6} (4)
      
                        (3) edge node [above, font=\small\bfseries] {0.4} (4)
      
                   (4) edge node [ right, font=\small\bfseries] {0.5} (5)
                   ;
                      
                    
    \end{tikzpicture}
    }
    \hspace{0.5cm}
     \subfloat[]{ \label{probgraph2}
      \begin{tikzpicture}[auto, node distance=2cm, every loop/.style={},
                    thick,main node/.style={scale=0.6, 
                    circle,draw,font=\sffamily\small\bfseries}]
                    \node[main node,fill={orange}] (1) {\textcolor{white}{1}};
                    \node[main node,fill={orange}] (2) [right of = 1] {\textcolor{white}{2}};
                     \node[main node,fill={orange}] (3) [below of = 1] {\textcolor{white}{3}};
                    \node[main node,fill={orange}] (4) [below of = 2] {\textcolor{white}{4}};
                      \path[every node/.style={scale=0.6,
                      font=\sffamily\small}]
                     
                      
                      (1) edge[] node [above, font=\small\bfseries] {0.4} (2)
                       edge[] node [pos = 0.5, left, font=\small\bfseries] {0.6} (3)
                        (2) edge[] node [right,pos=0.5, font=\small\bfseries] {0.6} (4)
      
                        (3) edge node [above, font=\small\bfseries] {0.4} (4)

                   ;
                      
                    
    \end{tikzpicture}
    }
    \caption{\color{black} a) Probabilistic graph $\G$, b) (2,0.2)-core $\mathcal{H}$ of $\G$.}
    \label{exam1}
\end{figure}
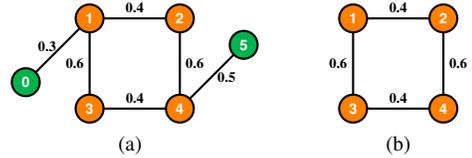

Let $u$ be a vertex in $\mathcal{G}$. The probability that $u$ has degree at least $t$ in $\G$ can be expressed as $\text{Pr}[ \textsf{deg}_{\G}(u) \geq t]=\sum_{G \sqsubseteq  \G }\text{Pr}(G) \cdot  \mathbbm{1}(G,u,t)$, where $\mathbbm{1}(G,u,t)$ is an indicator function which takes on 1 if degree of $u$ in possible world $G$ is at least $t$. It should be noted that as $t$ decreases (increases), $\text{Pr}[\textsf{deg}_{\G}(u) \geq t]$ increases (decreases). 
Given a user-defined threshed $\eta \in [0,1]$, the $\eta$-degree of $u$~\cite{bonchi2014core}, denoted by $\eta$-$\textsf{deg}_{\G}(u)$, is defined as the maximum integer $t \in [0,d_u]$ for which 
$\text{Pr}[ \textsf{deg}_{\G}(u) \geq t] \geq \eta$, where $d_u$ is the number of edges incident to $u$ which is equal to the deterministic degree of $u$.

\subsection{Core Decomposition in Probabilistic Graphs}
We use the notion of $(k,\eta)$-core in~\cite{bonchi2014core} for core decomposition in probabilistic graphs. Let $\mathcal{G}=(V,E,p)$ be a probabilistic graph, and $\eta \in [0,1]$ be a user-specified threshold. The $(k,\eta)$-\textit{core} is the largest subgraph $\mathcal{H}$ of $\mathcal{G}$ in which each vertex $u$ has $\eta$-degree no less than $k$, i.e. $\eta$-$\textsf{deg}_{\mathcal{H}}(u) \geq k$. 

{\em Core decomposition} of $\G$ is the set of all $(k,\eta)$-cores, for $k \in [0,k_{\max, \eta}]$, where $k_{\max,\eta}=\max_{u} \{ \eta$-$\textsf{deg}_{\G}(u) \}$. The {\em core number} of a vertex $u$, $\kappa_{\eta}(u)$, 
is the largest integer $k$ for which $u$ belongs to a $(k,\eta)$-core.

\begin{examp}
Consider Fig.~\ref{probgraph1}, vertex $u=1$, and $\eta = 0.2$. 
We have $\text{Pr}[ \textsf{deg}_{\G}(u) \geq 3] =  0.3 \cdot 0.4 \cdot 0.6 = 0.072$ 
(product of probabilities that edges $(0,1)$, $(1,2)$, and $(1,3)$ exist), and $\text{Pr}[ \textsf{deg}_{\G}(u) \geq 2] = 0.396$.
Since $0.396$ is greater than $\eta$, $\eta$-$\textsf{deg}_{\G}(u) = 2$.

\smallskip
Fig.~\ref{probgraph2} shows a $(2,0.2)$-core $\mathcal{H}$ of $\G$. Each vertex $u \in \mathcal{H}$, has $\eta$-degree $2$ with probability $0.24$. 

\smallskip
Consider $u=1$ and $\eta=0.2$.
Vertex $u$ is in $(1,0.2)$-core ($\G$ itself) and $(2,0.2)$-core ($\mathcal{H}$). There is no $(3,0.2)$-core, thus, $\kappa_{\eta}(u)=2$.
\end{examp}

\subsection{$\eta$-degree Computation Using Dynamic Programming (DP)}
We have $\text{Pr}[ \textsf{deg}_{\G}(u) \geq t] = \text{Pr}[ \textsf{deg}_{\G}(u) \geq t-1] - \text{Pr}[ \textsf{deg}_{\G}(u) = t]$.
To find $\eta$-degree of each vertex $u$, we need to compute 
$\text{Pr}[ \textsf{deg}_{\G}(u) = t]$. These probabilities can be computed using dynamic programming (DP) as proposed in~\cite{bonchi2014core}. 

The main idea of DP is as follows~\cite{bonchi2014core}. Given a vertex $u$ in probabilistic graph $\mathcal{G}$, and edge $e$ incident to $u$, the probability that $u$ has degree equal to $t$ ($\text{Pr}[ \textsf{deg}_{\G}(u) = t]$) consists of two mutually exclusive events: (1) edge $e$ exists and $u$ has degree $t-1$ in $\mathcal{G}_{\setminus \{ e \}}$, (2) edge $e$ does not exist and $u$ has degree $t$ in $\mathcal{G}_{\setminus \{ e \}}$, where $\mathcal{G}_{\setminus \{ e \}}$ is the subgraph of $\mathcal{G}$ in which edge $e$ does not exist. As a result, $\text{Pr}[ \textsf{deg}_{\G}(u) = t]$ can be written as the sum of the probability of these events, and a recursive formula is obtained.  
The above reasoning can be extended to any subgraph of $\mathcal{G}$. \cite{bonchi2014core} provides a thorough formulation of what is explained here.

\section{Related Work}
Core decomposition is one of the most popular notions of cohesive subgraphs~\cite{li2010computational,malliaros2020core,ugander2012structural}. It can be used for computing other definitions of dense subgraphs such as maximal cliques~\cite{eppstein2010listing}.
In deterministic graphs, core decomposition has been studied extensively in different settings~\cite{batagelj2003m,montresor2013distributed,wen2016efficient,aridhi2016distributed}. 

For probabilistic graphs, the notion of $(k,\eta)$-core is introduced by Bonchi \emph{et al.}~\cite{bonchi2014core}. The authors propose an algorithm that is based on iterative removing of the vertex of smallest $\eta$-degree, and updating the $\eta$-degree of its neighbours. In~\cite{bonchi2014core}, techniques are developed based on dynamic programming for computing $\eta$-degrees. More efficient algorithms are proposed by Esfahani \emph{et al.}~\cite{esfahani2019efficient} which can handle large graphs which do not fit in main memory. 
A different probabilistic core decomposition model, $(k,\theta)$-cores, is proposed by Peng \emph{et al.}~\cite{peng2018efficient} which is based on finding subgraphs whose vertices have a high probability to be a deterministic $k$-core member in different possible worlds of a probabilistic graph $\mathcal{G}$. 
Additionally, Yang \emph{et al.}~\cite{yang2019index} defined an index-based structure for processing core decomposition in probabilistic graphs.

Truss decomposition is another notion of dense substructures. For probabilistic graphs, the notion of local $(k,\eta)$-truss is introduced by Huang \emph{et al.} in \cite{huang2016truss}. The authors propose an algorithm for computing local $(k,\eta)$-truss which is based on iterative peeling of edges with support less than $k-2$, and updating the support of affected edges. Moreover, the notion of global $(k,\eta)$-truss is proposed in~\cite{huang2016truss} which is based on the probability of each edge belonging to a connected $k$-truss in a possible world. An approximate algorithm for the local truss decomposition is proposed by Esfahani \emph{et al.} in~\cite{esfahani2019fast} to efficiently compute the tail probability of edge supports in the peeling process described in~\cite{huang2016truss}.

\section{Proposed Approach}\label{sec:method}

As mentioned before, in M-PA we add two data screening stages before PA. The goal of the added data filtering steps is to reduce the number of vertices in the graph and speed up the follow-up analyses. In particular, we wish to remove a large proportion of low connectivity vertices (i.e. vertices with small $\eta$-degree) that we know will not likely be members of dense sub-communities in the graph.

\subsection{Data Screening Based on Degree Expectation}

Here we briefly explain the methodology behind the first stage of data screening. 
Given a probabilistic graph $\G=(V,E,p)$, for each vertex $v \in V$, we have a set of edges incident to $v$ and each edge is accompanied with a probability of existence $p_i$ that is independent of other edge probabilities in $\G$.

For vertex $v$, $\textsf{deg}_{\G}(v)$ can be interpreted as the sum of a set of independent Bernoulli random variables $X_i$'s with different success probabilities $p_i$'s \cite{esfahani2019efficient} where:

\begin{equation}
X_i = \begin{cases}
    1, & \text{if edge $e_i$ incident to $v$ exists in the graph}\\
    0, & \text{otherwise}
  \end{cases}
\end{equation}

and $\textsf{deg}_{\G}(v)$ follows Poisson binomial distribution with $E[\textsf{deg}_{\G}(v)]=\sum E[X_i]=\sum p_i$. We will use $\sum p_i$ as the first screening criteria since $\sum p_i$ can be seen as an approximation to $\textsf{deg}_{\G}(v)$. 

Thresholds are user-defined so any positive integer greater or equal to 0 is accepted. However, we recommend that the first threshold be set greater or equal to 5 (for example, if the first threshold is set to be 5, all vertices with degree expectation less than 5 is removed, and only those with $\sum p_i\ \ge\ 5$ and $\sum (1-p_i)\ \ge\ 5$ are kept). 
The purpose of this step is to screen out vertices that are rarely connected with others and hence are not likely to be part of any highly connected sub-network. For example, if a vertex $u$ has $\sum p_i$ less than 5, its $\textsf{deg}_{\G}(u)$ will also likely be around 5 with slight variations, therefore $u$ will not appear in cores with high activities (e.g. vertices with big coreness). Note that when the first threshold is set lower, more vertices will be retained. On the one hand, to speed up subsequent analyses, the threshold value should be high enough, on the other hand, the threshold should not be too high that possible highly connected vertices are removed. In our experiment, we empirically chose a conservative number, 5, as the default first threshold, but other threshold values could be used. 

\subsection{Data Screening Based on Lower-bounds of \textit{$\eta$-degree}}

In this section, we introduce the second data screening step before PA. For the remaining vertices that passed the first stage of data screening, we calculate lower-bounds of their $\eta$-degree using Lyapunov Central Limit Theorem (CLT)~\cite{Lyapunov-Nouvelle}. Given a vertex $v \in V$, based on Lyapunov CLT, $Z = \frac{1}{\sigma} \sum_{i=1}^{d_v} (X_i-\mu_i)$ has standard normal distribution, where $\mu_i = \Pr(X_i)$, and $ \sigma = \sqrt{\sum_{i=1}^{d_v}\Pr(X_i)(1-\Pr(X_i))}$. Approximation of $\text{Pr}[ \textsf{deg}_{\G}(v) \geq t] = \text{Pr}[ \sum_{i=1}^{d_v} X_i \geq t]$ can be obtained by subtracting $\mu_i$ from the sum of $X_i$'s, and dividing by $\sigma$. As a result, we have:
\begin{equation}\label{clt}
  \Pr\left[\sum_{i=1}^{d_v}X_i \geq t\right] =  \Pr\left[\frac{1}{\sigma} \sum_{i=1}^{d_v} (X_i - \mu_i) \geq \frac{1}{\sigma}\left(t- \sum_{i=1}^{d_v} \mu_i\right)\right]
\end{equation}
Since $Z$ has standard normal distribution, we can find the maximum value of $t$, such that the right-hand side of Equation~\ref{clt} is no less than $\eta$. 

We then use the second user-defined threshold to further select applicable vertices. The procedure for the second data screening stage is described in Algorithm~\ref{euclid}. Note that as we start graph peeling the vertex's $\eta$-degree will also start decreasing, so in this last data filtering stage, we only select based on vertices' initial $\eta$-degree lower-bounds.

\begin{algorithm} 
    \caption{Selection based on $\eta$-degree lower-bounds}\label{euclid}
    \begin{algorithmic}[1]
        \Procedure{SecondStageScreening ()}{}
        \State $\textit{nodelist} \gets \text{list of remaining nodes in }\textit{network}$
        \State $init\_\eta\_degree \gets \{\}$
        \Comment{empty hash table}
        \ForAll {$v \in \textit{nodelist}$}
            \State  $init\_\eta\_degree[v] \gets \text{compute initial }\eta\text{-}\textsf{deg}(v)$
        \EndFor
        \ForAll {$v \in init\_\eta\_degree.keys()$}
            \If {$init\_\eta\_degree[v] < \textit{threshold}$}
                \State delete $init\_\eta\_degree[v]$ \Comment{delete $v$ from hash table keys}
            \EndIf
        \EndFor
        \Return $init\_\eta\_degree.keys()$ \Comment{return hash table keys}
        \EndProcedure
    \end{algorithmic}
\end{algorithm}

\subsection{Remaining Parts of M-PA}

Core Decomposition which is based on peeling algorithm, includes three important steps: (1) removing vertex $u$ of the smallest $\eta$-degree, (2) assigning the core number of $u$ to be equal to its $\eta$-degree, and (3) recomputing the $\eta$-degree of $u$'s neighbours. Vertices should be kept sorted by their current $\eta$-degree at all times during the process. This process is challenging in probabilistic graphs as it involves many recomputations of $\eta$-degrees. It should be noted that computing $\eta$-degree of a vertex $u$ using dynamic programming takes $O(d_u^2)$. As a result, in~\cite{esfahani2019efficient}, an efficient version of the peeling algorithm is proposed which uses efficient array structures and lazy updates of $\eta$-degree of vertices. 

In M-PA, which is based on the PA proposed in~\cite{esfahani2019efficient}, we utilize data screening for detecting and removing non-promising vertices. 

The core computation part of M-PA approach is given in Algorithm~\ref{bz1}. Let $\text{V}_{\text{alive}}$ be the set of vertices which are remained after the second data screening phase, i.e. \textit{SecondStageScreening} (Line~\ref{valive}). The vertices are labelled by numbers 0 to $n-1$. Array \textbf{d} initially stores for each vertex the lower-bound on the $\eta$-degree of that vertex, and by the end of iterations we have the output core numbers in array \textbf{d}. The lower-bounds are obtained using CLT. Array \textbf{A} stores vertices in ascending order of their lower-bounds. 

Array \textbf{gone} keeps track of the removed vertices at each step of the algorithm. Array \textbf{valid} tells for each vertex $v$ if the $\eta$-degree of $v$ is the same as the value $\textbf{d}[v]$. These two arrays are initially set to all-false vectors (Lines~\ref{gone}-\ref{valid}) since all the vertices are on their lower-bounds at the beginning of the algorithm. Also, none of the vertices has been removed yet. 

The algorithm starts processing the vertices based on their (lower-bound on) $\eta$-degree. When a vertex $v$ is being processed, the algorithm checks if $v$ is on its lower-bound or its $\eta$-degree is available (Line~\ref{validitycheck}). if not, the $\eta$-degree of $v$ is computed using DP, stored in array \textbf{d}, and the vertex is swapped to be placed in a correct position in \textbf{A} (Lines~\ref{swapright1}-\ref{swapright2}). It should be noted that two additional arrays are defined to keep array \textbf{A} sorted at all times during the algorithm. One array stores the position of each vertex in \textbf{A}, and the other one stores the index boundaries of the vertex blocks having the same $\eta$-degree (exact or lower-bound) in \textbf{d} (a detailed discussion on the array structures can be found in~\cite{esfahani2019efficient}). These arrays help to swap vertex $v$ efficiently to its proper place in \textbf{A}. Otherwise, if $d[v]$ is the same as $\eta$-degree of $v$, 
$v$ is removed (Line~\ref{removed}), and the value $\textbf{d}[u]$ of all its neighbours $u \in  \text{V}_{\text{alive}}$ with $\textbf{d}[u] > \textbf{d}[v] $ is decremented by one (the same as deterministic graphs, in which the degree of a vertex decreases by one when a neighbour of that vertex is removed). Then, $u$ is swapped to a proper place in \textbf{A} (Lines~\ref{swapleft}-\ref{swapleft2}). Lines~\ref{extra1}-\ref{extra2} make sure that the algorithm does not go below the current minimum lower-bound which is being processed. It should be noted that during the main algorithm cycle, $\eta$-degree computation for each vertex is done with respect to its neighbours $u'$ such that $u' \in \text{V}_{\text{alive}}$, and \textbf{gone}$[u'] = \textit{false}$.

\noindent
The numerical stability, lower-bound accuracy, and correctness of the CLT-based peeling algorithm have been discussed extensively by Esfahani \emph{et al.} in~\cite{esfahani2019efficient} so we will not reiterate those aspects in this paper.

\begin{algorithm}
\caption{M-PA core decomposition function}\label{bz1}
\begin{algorithmic}[1]
    \Function{CoreCompute}{$\text{Graph}$ $\mathcal{G}$, $\eta$} 
    \State {$ \text{V}_{\text{alive}} \gets \text{SecondStageScreening ()}$} \label{valive}
    \State {{\textit{initialize}} \textbf{A}, and \textbf{d} based on $v \in \text{V}_{\text{alive}}$ }  
    \State \textbf{gone} $ \gets $ \textbf{False} \Comment{all-false vector} \label{gone}
    \State \textbf{valid} $ \gets $ \textbf{False} \Comment{all-false vector} \label{valid}
    \State $i \gets 0$
    \While {$i<n$}
    \State $v \gets \textbf{A}[i]$
            \If {\textbf{valid}$[v] = \textit{true}$} \label{validitycheck}
                \State \textbf{gone}$[v] \gets \textit{true}$ \label{removed}
                \ForAll  {$u: (u,v) \in \mathcal{N}_v$ and $ u \in \text{V}_{\text{alive}}$}
                     \If {$\textbf{d}[u] = \textbf{d}[v]$} \label{extra1}
                         \If {\textbf{valid}$[u] = \textit{false}$}
                                \State {Compute $\eta\text{-}\textsf{deg}(u)$,  \textbf{d}$[u] \gets \eta\text{-}\textsf{deg}(u)$}
                                 \State{swap $u$ to a correct place in \textbf{A} }
                          \EndIf \label{extra2}
                      \EndIf
                       \If {$\textbf{d}[u] > \textbf{d}[v]$}
                            \State {\textbf{d}$[u]--$, swap $u$ to a correct position in \textbf{A}} \label{swapleft}
                             \State{\textbf{valid}$[u] \gets \textit{false}$} \label{swapleft2}
                       \EndIf
                \EndFor
                \State $i++$
                \Else
                \State{\text{Compute} $\eta$-$\textsf{deg}(v)$, $ \textbf{d}[v] \gets \eta\text{-}\textsf{deg}(v)$} \label{swapright1}
                \State{\text{swap $v$ to a correct place in \textbf{A}}} \label{swapright2}
                \EndIf
                \EndWhile
            \State \textbf{return} $\textbf{d}$
    \EndFunction
\end{algorithmic}
\end{algorithm}

\section{Experiments}
In this section, we present results from the running time comparisons between M-PA approach and the original PA approach. We also compute the probabilistic density and probabilistic clustering coefficient for the outcomes of PA and M-PA for cohesiveness comparison. Both PA and M-PA are implemented in Java and the experiments are conducted using the WestGrid 
\footnote{\href{www.westgrid.ca}{www.westgrid.ca}} Graham cluster from Compute Canada \footnote{\href{www.computecanada.ca}{www.computecanada.ca}}.

\subsection{Efficiency Comparison}
We use the Flickr, DBLP, Biomine, and ljournal-2008 datasets used in \cite{esfahani2019efficient} and three more datasets (itwiki-2013, uk-2014-tpd, enwiki-2013) from Laboratory for Web Algorithmics (LAW)~\cite{BoVWFI,BRSLLP}. The dataset statistics are presented in Table \ref{table:01} and the description of each dataset is also given. The smallest dataset is Flickr with less than 30\ 000 vertices and the biggest dataset is ljournal-2008 with more than 5 million vertices and nearly 50 million edges.

\def\arraystretch{1.4}
\begin{table}[h]
\centering
\caption{Dataset statistics}
\begin{tabular}{cccc}
\hline
Name          & $|V|$       & $|E|$        & $P_{avg}$\\
\hline\hline
Flickr        & 24 125    & 300 836    & 0.13\\
\hline
DBLP          & 684 911   & 2 284 991  & 0.26\\
\hline
Biomine       & 1 008 201 & 6 722 503  & 0.27\\
\hline
itwiki-2013   & 1 016 867 &	23 429 644 & 0.50\\
\hline
uk-2014-tpd   & 1 766 010 & 15 283 718 & 0.50\\
\hline
enwiki-2013   & 4 206 785 & 91 939 728 & 0.50\\
\hline
ljournal-2008 & 5 363 260 & 49 514 271 & 0.50\\  
\hline
\end{tabular}
\label{table:01}
\end{table}

\begin{itemize}
    \item \textbf{Flickr}: snapshot of the Flickr online photo sharing community. The edge probability between any two nodes (users) is computed based on the Jaccard coefficient of the groups the users belonged to~\cite{bonchi2014core}.
    \item \textbf{DBLP}: snapshot of the DBLP database. Two authors (nodes) are linked if they have coauthored a publication together and the edge probability is computed based on the number of collaborations~\cite{bonchi2014core}.
    \item \textbf{Biomine}: snapshot of the Biomine probabilistic database. Biomine integrates indexes from several biological databases (Entrez Gene, STRING, UniProt, etc.) and probability is calculated for all edges (i.e. cross-references)~\cite{eronen2012biomine}.
    \item \textbf{itwiki-2013}, \textbf{enwiki-2013}, \textbf{uk-2014-tpd}: snapshots of the Italian and English part of Wikipedia in 2013, and the snapshot of top private .uk domains in 2014~\cite{BoVWFI,BRSLLP}. We generated $[0, 1]$  uniformly distributed probabilities for the edges.
    \item \textbf{ljournal-2008}: snapshot of LiveJournal social network in 2008, each node is a user and an edge from node $x$ to node $y$ indicates $x$ registered $y$ as its friend~\cite{BoVWFI,BRSLLP}. We generated probability values uniformly distributed in $[0, 1]$ for the edges.
\end{itemize}

For each dataset, we record the running time for PA and M-PA separately. In order to avoid our benchmark task compete for memory bandwidth with other jobs on the cluster, we requested an entire 32-core compute node on Graham with two Intel(R) Xeon(R) E5-2683v4 CPU@2.1GHz, and 125GB RAM. In addition, we have also set up the input and output of the algorithm to communicate directly with the compute node to avoid the impact from the parallel file system used in Graham's login node on the benchmark results.

As discussed before, M-PA takes the same arguments as the original PA and two more user-defined thresholds for data screening ($threshold_1$, $threshold_2$). The first threshold is not affected by the choice of $\eta$, but the second threshold is related to $\eta$ because it is for the initial $\eta$-degree. There are many possible threshold selection methods and in some cases in the actual analysis, according to the specific dataset or purpose, it may be necessary to often change the chosen threshold for better outcomes. In this section, we set $\eta$ to be 5 different values: 0.1, 0.3, 0.5, 0.7, and 0.9. For each dataset and $\eta$, we performed exploratory analyses to determine the data screening thresholds. 

For the first threshold, we calculate the $\sum p_i$ of all vertices. If the results' 75th percentile is less than or equal to 5, we use 5 as the threshold for the first data filtering step. Otherwise, we assume that the distribution of $\sum p_i$ has an inflection point where the value of $\sum p_i$ quickly grows, and we use piecewise regression (segmented regression) to detect this change point and set it to be the first threshold. If more than one infection point is discovered, the highest one will be used. We have explained the rationale to choose number 5 as the default first threshold in Section~\ref{sec:method}: we wish to remove low-connectivity nodes in the network that are not eligible to be part of any highly connected dense subgraphs but at the same time the default threshold should not be too high that possible valuable information is lost. For the determination of the second data screening threshold, for convenience we assume all the vertices have passed the screening of the first step; then we calculate the initial $\eta$-degree of the current list of vertices in the graph. If more than 80\% of the results (80th percentile of the initial $\eta$-degrees for current vertices in the graph) are less than or equal to 10, then we choose 10 as the threshold for the second data screening step; otherwise, we still apply segmented regression to locate the second threshold. The reason to choose the arbitrary number 10 as the default second data screening threshold is that if a vertex has at least 10 edges incident to it before peeling, we can consider it a hotspot suited for the afterward high activity subgraph mining. If in the full graph a vertex is not connected to at least 10 other vertices, there is no point in performing core decomposition as we only focus on dense sub-communities.

Note that $\eta$-degree is related to the choice of $\eta$. Typically, the higher the $\eta$, the lower the $\eta$-degree. For example, if we set different $\eta$ to M-PA with the same dataset and same two-stage screening thresholds, higher $\eta$ would result in more nodes being screened out and this should also be included in the consideration when determining the data screening thresholds. Ultimately the choice of threshold combination would be depending on the applications, e.g. if the volume of the dataset is too big to reason and we wish to reduce its size significantly, then setting higher $\eta$ (0.5, 0.7, 0.9) and higher data-screening thresholds would certainly help.

As a result, the thresholds we obtained for our experiment are presented in Table \ref{table:02}. The majority of data screening threshold combinations is $(5,10)$ and this is due to the distribution of edge existing probabilities in the datasets that we used. In addition, using different threshold selection methods will result in different threshold combinations to ours.

\def\arraystretch{1.4} 
\begin{table*}
\centering
\caption{Thresholds for data screening}
\begin{tabular}{ccccccccc}
\hline
 & $\eta$   & Filckr & DBLP & Biomine & itwiki-2013 & uk-2014-tpd & enwiki-2013 & ljournal-2008 \\
\hline\hline
\multicolumn{2}{l}{$threshold_1$} & 5      & 5    & 5  & 41 & 5 & 30 & 21\\
\hline
\multirow{5}{*}{$threshold_2$} & 0.1 & 10 & 10 & 10 & 39 & 10 & 34 & 14\\
 & 0.3 & 10     & 10   & 10 & 36 & 10 & 32 & 10 \\
 & 0.5 & 10     & 10   & 10 & 36 & 10 & 30 & 10 \\
 & 0.7 & 10     & 10   & 10 & 34 & 10 & 29 & 10 \\
 & 0.9 & 10     & 10   & 10 & 32 & 10 & 27 & 10 \\
\hline
\end{tabular}
\label{table:02}
\end{table*}

\begin{figure}[ht]
\begin{center}
\includegraphics[scale=0.55]{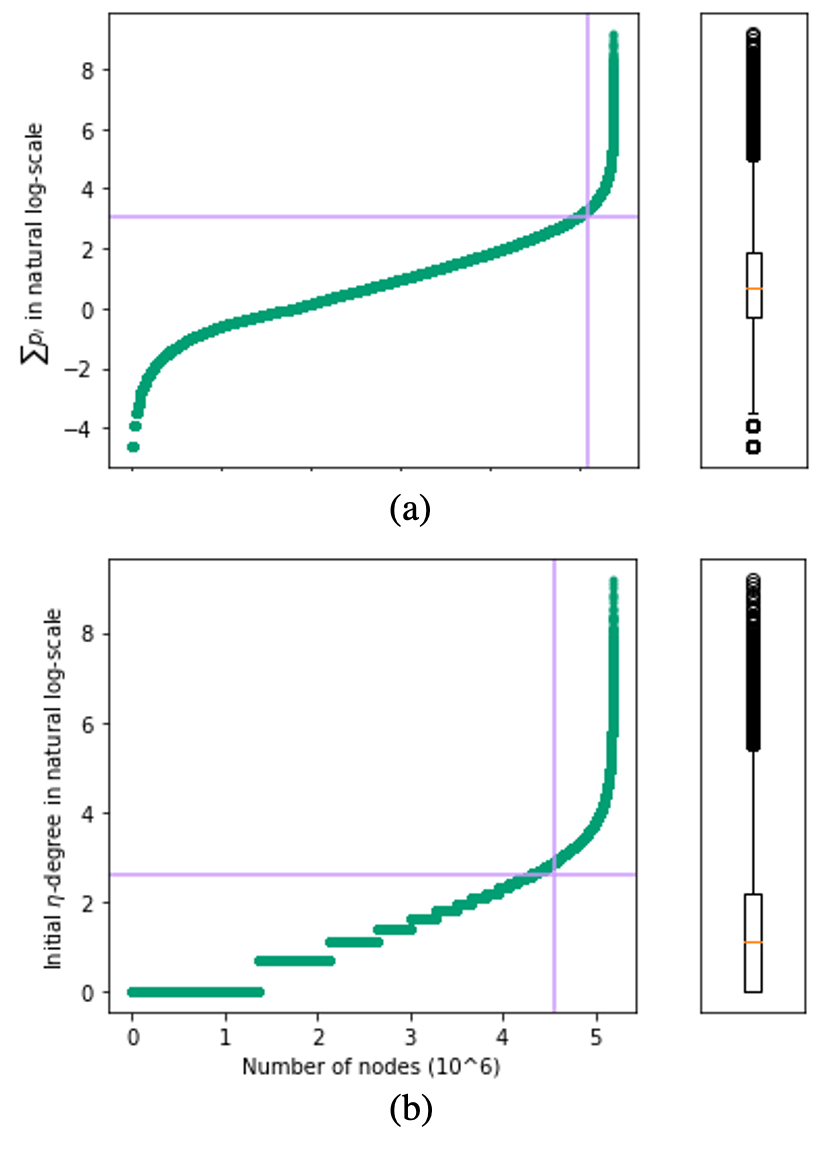}
\end{center}
\caption{a) Distribution of $\sum p_i$ for ljournal-2008, b) Distribution of initial $\eta$-degree for ljournal-2008 with $\eta=0.1$.}
\label{fig2}
\end{figure}

For itwiki-2013, uk-2014-tpd, enwiki-2013, and ljournal-2008, the threshold combinations start to vary, and we will use the case of ljournal-2008 as an example. For ljournal-2008, the third quartile of $\sum p_i$ is 6.52 ($\approx 1.87$ in log-scale, as shown in Fig.~\ref{fig2}a), which is greater than 5. So we performed segmented regression and found 21 ($\approx 3.04$ in log-scale) to be the first threshold, as illustrated in Fig.~\ref{fig2}a. Piecewise regression was also applied on the initial $\eta$-degree result of ljournal-2008 with $\eta=0.1$, since its 80th percentile, 11, is greater than 10. As shown in Fig.~\ref{fig2}b, we found 14 ($\approx 2.64$ in log-scale) to be the second threshold for $\eta=0.1$ case of ljournal-2008 dataset.

\begin{figure*}
    \centering
    \subfloat{ 
    \begin{tikzpicture}
\begin{axis}[width=3.8cm,height=3.5cm,
xtick pos=left,
ytick pos=left,
    title={\textbf{Flickr}},
    xlabel={\textbf{$\eta$}},
     xlabel style={font=\fontsize{7}{7}\selectfont},
   symbolic x coords={0.1,0.3,0.5,0.7,0.9},
    xticklabel style={rotate=-45},
    ytick distance =1,
    xtick={0.1,0.3,0.5,0.7,0.9},
    ylabel={\textbf{Time (s)}},
    ylabel near ticks,
    ticklabel style = {font=\fontsize{7}{6}\selectfont},
    legend style={draw=none,nodes={scale=0.7}},
]

\addplot[
    color=blue,
    mark=square,
    ]
    coordinates {
    (0.1,6)(0.3,4)(0.5,4)(0.7,5)(0.9,3)
    };
    \addplot[
    color=red,
    mark=*,
    ]
    coordinates {
    (0.1,3)(0.3,3)(0.5,2)(0.7,2)(0.9,2)
    };
    
\end{axis}
\end{tikzpicture}
    }
\hspace{0.01cm}
\subfloat{ 
    \begin{tikzpicture}
\begin{axis}[width=3.8cm,height=3.5cm,
xtick pos=left,
ytick pos=left,
    title={\textbf{DBLP}},
    xlabel={\textbf{$\eta$}},
     xlabel style={font=\fontsize{7}{7}\selectfont},
    symbolic x coords={0.1,0.3,0.5,0.7,0.9},
    xticklabel style={rotate=-45},
    xtick={0.1,0.3,0.5,0.7,0.9},
    ylabel near ticks,
    ticklabel style = {font=\fontsize{7}{6}\selectfont},
    legend style={draw=none,nodes={scale=0.7}},
]

\addplot[
    color=blue,
    mark=square,
    ]
    coordinates {
    (0.1,8)(0.3,8)(0.5,8)(0.7,8)(0.9,8)
    };
    \addplot[
    color=red,
    mark=*,
    ]
    coordinates {
    (0.1,3)(0.3,3)(0.5,2)(0.7,3)(0.9,3)
    };
    
\end{axis}
\end{tikzpicture}
    }
\hspace{0.01cm}
\subfloat{ 
    \begin{tikzpicture}
\begin{axis}[width=3.8cm,height=3.5cm,
xtick pos=left,
ytick pos=left,
   title={\textbf{Biomine}},
    xlabel={\textbf{$\eta$}},
     xlabel style={font=\fontsize{7}{7}\selectfont},
   ymin=50,
  ytick distance =2,
  symbolic x coords={0.1,0.3,0.5,0.7,0.9},
    xticklabel style={rotate=-45},
    ticklabel style = {font=\fontsize{7}{6}\selectfont},
    xtick={0.1,0.3,0.5,0.7,0.9},
    ylabel near ticks,
    legend style={draw=none,nodes={scale=0.7}},
]

\addplot[
    color=blue,
    mark=square,
    ]
    coordinates {
    (0.1,53)(0.3,55)(0.5,56)(0.7,58)(0.9,60)
    };
    \addplot[
    color=red,
    mark=*,
    ]
    coordinates {
   (0.1,53)(0.3,55)(0.5,54)(0.7,56)(0.9,52)
    };
    
\end{axis}
\end{tikzpicture}
   }
\hspace{0.01cm}
\subfloat{ 
    \begin{tikzpicture}
\begin{axis}[width=3.8cm,height=3.5cm,
xtick pos=left,
ytick pos=left,
    title={\textbf{itwiki-2013}},
    xlabel={\textbf{$\eta$}},
     xlabel style={font=\fontsize{7}{7}\selectfont},
    symbolic x coords={0.1,0.3,0.5,0.7,0.9},
    xticklabel style={rotate=-45},
    xtick={0.1,0.3,0.5,0.7,0.9},
    ylabel near ticks,
    ticklabel style = {font=\fontsize{7}{6}\selectfont},
    legend style={draw=none,nodes={scale=0.7}},
]

\addplot[
    color=blue,
    mark=square,
    ]
    coordinates {
    (0.1,97)(0.3,98)(0.5,99)(0.7,98)(0.9,102)
    };
    \addplot[
    color=red,
    mark=*,
    ]
    coordinates {
    (0.1,63)(0.3,64)(0.5,64)(0.7,62)(0.9,62)
    };
    
\end{axis}
\end{tikzpicture}
}

\hspace{0.01cm}
\subfloat{ 
    \begin{tikzpicture}
\begin{axis}[width=3.8cm,height=3.5cm,
xtick pos=left,
ytick pos=left,
    title={\textbf{uk-2014-tpd}},
    xlabel={\textbf{$\eta$}},
     xlabel style={font=\fontsize{7}{7}\selectfont},
    symbolic x coords={0.1,0.3,0.5,0.7,0.9},
    xticklabel style={rotate=-45},
    xtick={0.1,0.3,0.5,0.7,0.9},
    ylabel={\textbf{Time (s)}},
    ylabel near ticks,
    ticklabel style = {font=\fontsize{7}{6}\selectfont},
    legend style={draw=none,nodes={scale=0.7}},
]

\addplot[
    color=blue,
    mark=square,
    ]
    coordinates {
    (0.1,81)(0.3,79)(0.5,81)(0.7,81)(0.9,81)
    };
    \addplot[
    color=red,
    mark=*,
    ]
    coordinates {
    (0.1,77)(0.3,77)(0.5,76)(0.7,75)(0.9,75)
    };
    
\end{axis}
\end{tikzpicture}
}
\hspace{0.01cm}
\subfloat{ 
    \begin{tikzpicture}
\begin{axis}[width=3.8cm,height=3.5cm,
xtick pos=left,
ytick pos=left,
    title={\textbf{enwiki-2013}},
    xlabel={\textbf{$\eta$}},
     xlabel style={font=\fontsize{7}{7}\selectfont},
    symbolic x coords={0.1,0.3,0.5,0.7,0.9},
    xticklabel style={rotate=-45},
    xtick={0.1,0.3,0.5,0.7,0.9},
    ylabel near ticks,
    ticklabel style = {font=\fontsize{7}{6}\selectfont},
    legend style={draw=none,nodes={scale=0.7}},
]

\addplot[
    color=blue,
    mark=square,
    ]
    coordinates {
    (0.1,351)(0.3,336)(0.5,357)(0.7,358)(0.9,355)
    };
    \addplot[
    color=red,
    mark=*,
    ]
    coordinates {
    (0.1,233)(0.3,236)(0.5,224)(0.7,226)(0.9,228)
    };
    
\end{axis}
\end{tikzpicture}
}
\hspace{0.01cm}
\subfloat{ 
\begin{tikzpicture}
\begin{axis}[width=3.8cm,height=3.5cm,
legend pos=outer north east,
xtick pos=left,
ytick pos=left,
   title={\textbf{ljournal-2008}},
 xlabel={\textbf{$\eta$}},
  xlabel style={font=\fontsize{7}{7}\selectfont},
    xticklabel style={rotate=-45},
     xtick={0.1,0.3,0.5,0.7,0.9},
    ticklabel style = {font=\fontsize{7}{6}\selectfont},
    ylabel near ticks,
  legend entries={\textbf{PA},\textbf{M-PA}},
    legend style={font=\fontsize{7.5}{6}\selectfont,draw=none,nodes={scale=0.6}},
]

\addplot[
    color=blue,
    mark=square,
    ]
    coordinates {
    (0.1,158)(0.3,166)(0.5,168)(0.7,170)(0.9,178)
    };
    \addplot[
    color=red,
    mark=*,
    ]
    coordinates {
   
  (0.1,116)(0.3,122)(0.5,121)(0.7,121)(0.9,118)
    };
    
\end{axis}
\end{tikzpicture}
   }
    \caption{Running time of probabilistic core decomposition: PA vs M-PA.}
    \label{runningtime}
\end{figure*}
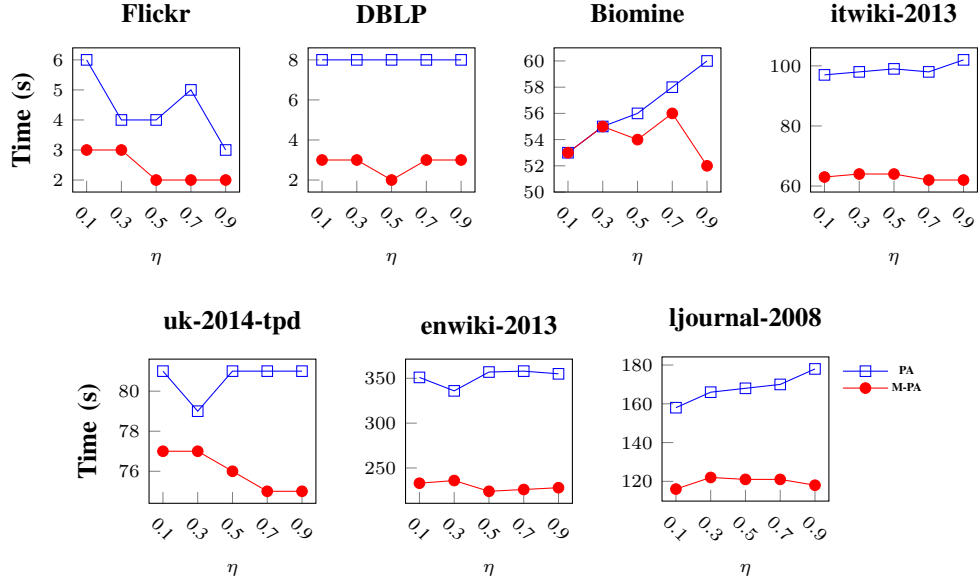

With the two extra threshold arguments located for M-PA, we conducted algorithm efficiency experiments and the results are illustrated in Fig.~\ref{runningtime}. It can be seen that in an overwhelming majority of cases, M-PA is faster than PA. The worst cases are for Biomine with $\eta=0.1$ and $\eta=0.3$, M-PA finished computation at the same time as PA. This is possibly due to the choice of screening thresholds. To ensure fair comparisons we only used a fixed method for threshold exploration. As stated before, in real cases for different datasets and $\eta$ we might need to adjust the screening thresholds using different methods in order to achieve optimal results for M-PA.

\subsection{Quality Evaluation}

In this section, we evaluate results from PA and M-PA in terms of graph cohesiveness. The metrics we used are probabilistic density (PD) and probabilistic clustering coefficient (PCC)~\cite{huang2016truss}. These metrics are defined as follows:

\begin{equation}
\mathrm{PD}(\mathcal{G})=\frac{\sum_{e \in E} p_e}{\frac{1}{2}|V| \cdot(|V|-1)}
\end{equation}

\begin{equation}
\operatorname{PCC}(\mathcal{G})=\frac{3 \sum_{\Delta_{u v w} \in \mathcal{G}} p(u, v) \cdot p(v, w) \cdot p(u, w)}{\sum_{(u, v),(u, w), v \neq w} p(u, v) \cdot p(u, w)}
\end{equation}

Simply put, PD is the sum of all edge probabilities in the graph divided by the maximum possible number of edges in the graph. PCC on the other hand is the degree measurement for nodes in the graph to cluster together.

We use $\eta=0.1$, $\eta=0.5$, and $\eta=0.9$ and we report the PD, PCC results for the maximum core (i.e. the densest subgraph) obtained from running PA and M-PA on Flickr, DBLP, Biomine, itwiki-2013, uk-2014-tpd, enwiki-2013, and ljournal-2008 in Table~\ref{table:03}. Sometimes for the given core number, we might discover several connected components in the results, so we report the average PD, PCC instead of the maximum.

\definecolor{Blue1}{HTML}{2e1f96} 
\definecolor{Blue2}{HTML}{6d62b6} 
\definecolor{Blue3}{HTML}{978fcb} 
\definecolor{Blue4}{HTML}{d5d2ea} 

\def\arraystretch{1.4}
\begin{table*}
\vspace{0.3cm}
\centering
\caption{Cohesiveness statistics from the original PA, O, and M-PA, M on Flickr, DBLP, Biomine, itwiki-2013, uk-2014-tpd, enwiki-2013, and ljournal-2008}
\begin{tabular}{ccccc}
\hline
Graph & $\eta$ & $Coreness_{O\_max}$/$Coreness_{M\_max}$ & $PD_{O\_avg}$ /$PD_{M\_avg}$ & $PCC_{O\_avg}$/$PCC_{M\_avg}$ \\
\hline\hline
Flickr   & 0.1 & 46/27   & \cellcolor{Blue4!95}1.0/0.871    & \cellcolor{Blue4!95}1.0/0.872     \\
         & 0.5 & 46/25   & \cellcolor{Blue4!95}1.0/0.871    & \cellcolor{Blue4!95}1.0/0.872     \\
         & 0.9 & 46/23   & \cellcolor{Blue4!95}1.0/0.871    & \cellcolor{Blue4!95}1.0/0.872     \\
DBLP     & 0.1 & 26/26   & \cellcolor{Blue1!46}0.264/0.264 & \cellcolor{Blue1!46}0.317/0.317 \\
         & 0.5 & 21/21   & \cellcolor{Blue1!46}0.264/0.264 & \cellcolor{Blue1!46}0.317/0.317 \\
         & 0.9 & 16/16   & \cellcolor{Blue1!46}0.419/0.419 & \cellcolor{Blue1!46}0.441/0.441 \\
Biomine  & 0.1 & 79/79   & \cellcolor{Blue1!46}0.212/0.212 & \cellcolor{Blue1!46}0.218/0.218 \\
         & 0.5 & 70/70   & \cellcolor{Blue1!46}0.227/0.227   & \cellcolor{Blue1!46}0.230/0.230 \\
         & 0.9 & 60/60   & \cellcolor{Blue1!46}0.216/0.216 & \cellcolor{Blue1!46}0.221/0.221 \\
itwiki-2013 & 0.1 & 118/117 & \cellcolor{Blue2!46}0.203/0.202 & \cellcolor{Blue4!95}0.035/0.031 \\
         & 0.5 & 110/108 & \cellcolor{Blue2!46}0.202/0.199 & \cellcolor{Blue4!95}0.036/0.031 \\
         & 0.9 & 102/101 & \cellcolor{Blue2!46}0.203/0.202 & \cellcolor{Blue4!95}0.035/0.031 \\
uk-2014-tpd & 0.1 & 257/257 & \cellcolor{Blue1!46}0.359/0.359 & \cellcolor{Blue1!46}0.361/0.361 \\
         & 0.5 & 244/244 & \cellcolor{Blue1!46}0.359/0.359 & \cellcolor{Blue1!46}0.361/0.361 \\
         & 0.9 & 231/231 & \cellcolor{Blue1!46}0.359/0.359 & \cellcolor{Blue1!46}0.361/0.361 \\
enwiki-2013 & 0.1 & 78/77 & \cellcolor{Blue3!46}0.122/0.111 & \cellcolor{Blue3!46}0.122/0.114 \\
         & 0.5 & 70/70 & \cellcolor{Blue2!46}0.112/0.113 & \cellcolor{Blue2!46}0.115/0.116 \\
         & 0.9 & 62/62 & \cellcolor{Blue2!46}0.110/0.111 & \cellcolor{Blue2!46}0.113/0.114 \\
ljournal-2008 & 0.1 & 156/156 & \cellcolor{Blue1!46}0.375/0.375 & \cellcolor{Blue1!46}0.378/0.378 \\
         & 0.5 & 147/147 & \cellcolor{Blue1!46}0.379/0.379 & \cellcolor{Blue1!46}0.381/0.381 \\
         & 0.9 & 138/138 & \cellcolor{Blue1!46}0.379/0.379 & \cellcolor{Blue1!46}0.381/0.381 \\
\hline
\end{tabular}
\label{table:03}
\begin{tabular}{cccccccc}
\textcolor{Blue1!46}{$\blacksquare$} & identical &
\textcolor{Blue2!46}{$\blacksquare$} & within 2\% change&
\textcolor{Blue3!46}{$\blacksquare$} & within 10\% change&
\textcolor{Blue4!88}{$\blacksquare$} & within 15\% change
\end{tabular}
\vspace{0.3cm}
\end{table*}

Table~\ref{table:03} is coloured based on levels of changes in PD and PCC results between PA and M-PA, i.e. if PD/PCC results are identical for PA and M-PA, or within 2\%, 10\%, or 15\% level of changes. It can be seen that M-PA produce very close if not identical PD/PCC results to the PA. The maximum level of changes is within 15\% and for DBLP, Biomine, uk-2014-tpd, and ljournal-2008, PA and M-PA produced identical PD and PCC scores. This is expected since our final goal underlying the modification to the original peeling algorithm is to make it focus more on dense cores and remove the nodes that do not belong to them. Ideally, when we selected and removed nodes below the user-defined thresholds, the dense cores in the graph should not be affected and the algorithm should run faster. In the case of Flickr, itwiki-2013, and enwiki-2013, M-PA gives slightly different PD and PCC scores, e.g., for $\eta=0.5$ case of Flickr, PA's maximum core is 12.9\% denser than M-PA, for $\eta=0.5$ case of enwiki-2013, M-PA's maximum core is 0.89\% denser than PA, etc. 
Flickr is the smallest dataset of the seven and has the smallest average edge existing probability. Given this, the thresholds we used might be too strict for the Flickr dataset. However, 0.87 is still very good as PD and PCC scores and having nearly 0.9 density is completely acceptable. As for itwiki-2013, the PD score is nearly identical (within 2\% change) between PA and M-PA and the PCC score is also very close. However, the PCC score is extremely low, which could indicate that for this specific dataset, the nodes do not tend to cluster together and hence the slightly more different PCC scores. Lastly, for enwiki-2013, at $\eta=0.1$ M-PA produced slightly small maximum coreness but the PD/PCC scores are still close to PA. Additionally, at $\eta=0.5$ and $\eta=0.9$, M-PA was able to produce a denser subgraph compared to PA.

\section{Conclusion}
We presented a multi-stage probabilistic graph peeling algorithm (M-PA) for core decomposition. A two-stage data filtering procedure was added to the original peeling algorithm (PA) to reduce the complexity of input graphs and increase the algorithm efficiency. We compared M-PA and PA in terms of speed and we showed that M-PA is generally faster than PA. After evaluating the cohesiveness from the results of M-PA and PA, we concluded that M-PA, when equipped with proper data screening thresholds, will produce very comparable if not identical subgraph density as the original PA, and at the same time will be more efficient.

\balance

\section*{Acknowledgment}

This work was enabled in part by support provided by WestGrid and Compute Canada.

\bibliographystyle{ieeetr}
\bibliography{reference}

\end{document}